\documentclass[11pt]{article}

\usepackage[final]{acl}

\usepackage{times}
\usepackage{latexsym}
\usepackage{graphicx}
\usepackage{amsmath}
\usepackage{booktabs}
\usepackage{float}
\usepackage[T1]{fontenc}
\usepackage[utf8]{inputenc}
\usepackage{microtype}
\usepackage{inconsolata}
\title{Evaluating Dedicated Monolingual and Joint Multilingual Causal Models for Dravidian Languages}

\author{
  Venkata Naga Sai Vishnu Rohit Pulipaka \\
  Independent Researcher \\
  \texttt{pvnsvrohit@gmail.com}
}

\begin{document}
\maketitle

\begin{abstract}
Dravidian languages, mainly Tamil, Telugu, Kannada, and Malayalam make up only a small part of the data used to train multilingual language models, so it's not clear how much per-language ability these models actually keep. I have trained five GPT-2 architecture models from scratch to compare four monolingual models (one each for Tamil, Telugu, Kannada, and Malayalam, each with its own 32K-vocabulary subword tokenizer) against one multilingual model sharing a 64K-vocabulary subword tokenizer across all four languages. All the 5 models are trained on cleaned CC-100, Wikipedia, and Samanantar data. I have tested the models on perplexity, bits-per-byte, tokenizer efficiency, and fine-tuning results which are compared against mGPT. The monolingual models outperform mGPT on sentiment classification and named entity recognition, and their tokenizers proved more efficient than the shared multilingual model across all the languages tested. 
\end{abstract}

\section{Introduction}
Tamil, Telugu, Kannada, and Malayalam are the major Dravidian languages which are agglutinative, morphologically complex, and written in their own Brahmic scripts, quite different from most languages studied in NLP. Despite long literary histories and a large online presence, they get far less NLP research attention than Latin script, high resource languages. Massively multilingual models do include Dravidian text, but per language capacity gets spread thin across hundreds of languages, and the shared tokenizers are usually built around Romanized text.

Large multilingual models like mBERT \cite{devlin-etal-2019-bert}, XLM-R \cite{conneau-etal-2020-unsupervised}, mT5 \cite{xue-etal-2021-mt5}, and mGPT \cite{shliazhko-etal-2024-mgpt} do include Dravidian languages, but as a small piece of mixtures spanning dozens or hundreds of languages at once. So how much capacity is actually left per language after that kind of dilution? And do these languages pay a bigger cost from a shared, general purpose tokenizer given how morphologically dense they are? Toolkits built specifically for Indian languages, IndicBERT \cite{kakwani-etal-2020-indicnlpsuite} and IndicTrans2 \cite{gala2023indictrans2} among them, have moved things forward quite a bit, but they mostly lean encoder focused or are built around translation objectives. Neither is designed to isolate what a from scratch, language dedicated causal model and tokenizer can actually do for this family.

I address this gap directly rather than through a multilingual proxy and  trained 5 GPT-2-architecture causal language models from scratch: 4 monolingual models, 1 per language, each paired with a dedicated 32K-vocabulary SentencePiece tokenizer trained only on that language's data, and one multilingual model sharing a 64K-vocabulary joint tokenizer across all four. Training data is drawn from curated CC-100, Wikipedia, and AI4Bharat Samanantar corpora.

\section{Related Work}
Models like mBERT \cite{devlin-etal-2019-bert}, XLM-R \cite{conneau-etal-2020-unsupervised}, mT5 \cite{xue-etal-2021-mt5}, mGPT \cite{shliazhko-etal-2024-mgpt}, and BLOOM \cite{JMLR:v25:23-0581} include Dravidian languages among dozens to hundreds of training languages. These models are general-purpose baselines and are the default choice for low-resource languages lacking dedicated pretraining, but their tokenizers and parameter budgets are allocated jointly across the full language mixture. Prior work has shown this allocation is non-uniform as high-resource languages receive un-evenly efficient tokenization and larger effective training signal \cite{petrov2023language, rust-etal-2021-good}. I use mGPT for multilingual baseline for downstream comparison.

AI4Bharat's IndicNLPSuite \cite{kakwani-etal-2020-indicnlpsuite} and IndicCorp provide tokenization, normalization, and corpus resources spanning Indian languages, while IndicBERT \cite{kakwani-etal-2020-indicnlpsuite} and the IndicGLUE benchmark established encoder-based evaluation suites, later extended with IndicXNLI \cite{aggarwal-etal-2022-indicxnli}. Google's MuRIL \cite{khanuja2021murilmultilingualrepresentationsindian} targets 17 Indian languages with a BERT-style encoder. IndicTrans2 \cite{gala2023indictrans2} addresses translation across 22 scheduled languages. These efforts substantially improved Indic NLP tooling, but are predominantly encoder-focused rather than causal decoder-only models, and treat Indian languages collectively rather than isolating the Dravidian family specifically. Similarly, modern generative suites like Sarvam demonstrate the value of regional pretraining and custom Indic tokenization, though they scale across dozens of language families simultaneously at multi-billion parameter regimes rather than isolating the Dravidian family under matched, lightweight bounds.

Prior work has demonstrated that monolingual models trained from scratch with dedicated tokenizers often outperform multilingual counterparts on specific target languages such as CamemBERT \cite{martin-etal-2020-camembert} for French, FinBERT \cite{virtanen2019multilingualenoughbertfinnish} for Finnish, and AraBERT \cite{antoun-etal-2020-arabert} for Arabic. This effect is attributed largely to tokenizer fertility: general-purpose multilingual vocabularies over-segment morphologically rich, non-Latin-script languages relative to a dedicated tokenizer, inflating sequence lengths and diluting context windows. Dravidian languages, being agglutinative and written in distinct Brahmic scripts, are ideal candidates for evaluating this effect under controlled, matched architectures.

\section{Methodology}

\subsection{Training Data}
For each of the 4 target languages, I collect text from 3 primary sources: CC-100 \cite{conneau-etal-2020-unsupervised}, Wikipedia (2023-11-01 dump), and the target side of AI4Bharat Samanantar \cite{ramesh-etal-2022-samanantar}. Kannada and Malayalam additionally include a filtered simple-narrative corpus. Table~\ref{tab:datastats} reports corpus size by split for each language.
\begin{table}[H]
\centering
\small
\begin{tabular}{lccc}
\toprule
\textbf{Language} & \textbf{Train} & \textbf{Val} & \textbf{Test} \\
\midrule
Telugu    & 312.9M & 3.49M & 3.50M \\
Tamil     & 646.8M & 12.09M & 12.10M \\
Kannada   & 209.8M & 2.46M & 2.46M \\
Malayalam & 364.3M & 4.21M & 4.21M \\
\bottomrule
\end{tabular}
\caption{Corpus size by split, in words (whitespace-delimited).}
\label{tab:datastats}
\end{table}
Raw text is Unicode-NFC normalized, stripped of HTML tags and URLs, and whitespace-collapsed; no deduplication is applied beyond source-level line filtering. Each language's cleaned corpus is split 96/2/2 into train/validation/test splits by contiguous line index to preserve structural coherence.

\subsection{Tokenization}
I trained one SentencePiece BPE tokenizer per language with 32,000 vocabulary size and one shared tokenizer across all 4 languages with 64,000 vocabulary size for the multilingual model. I used \texttt{nmt\_nfkc} normalization, full character coverage ($1.0$), and byte-level fallback to guarantee closed vocabulary coverage. Special token IDs (\texttt{<pad>}=0, \texttt{<unk>}=1, \texttt{<s>}=2, \texttt{</s>}=3) are fixed at training time. 

\subsection{Model Architecture and Pretraining}
All models follow a GPT-2-small-scale decoder-only architecture: 12 layers, 768 hidden size, 12 attention heads, 3072-dimensional feed-forward layers, GELU activation, and a 1024-token context window. This configuration yields 110M parameters for each monolingual model and 135M parameters for the multilingual model. All the models are trained only one time.

Each corpus line is tokenized independently and truncated to 1024 tokens. Training is conducted using the Hugging Face Trainer with an effective batch size of 256 sequences (per-device batch size $4 \times$ gradient accumulation $16 \times 4$ GPUs) for 3 epochs. I used AdamW with learning rate $1\text{e-}4$, cosine learning rate decay, 4,000 warmup steps, weight decay $0.01$, and gradient clipping at norm $0.5$ in \texttt{bf16}/\texttt{tf32} mixed precision with gradient checkpointing. Training runs on 4 GPUs via Hugging Face Accelerate on a SLURM cluster.

\subsection{Evaluation}
I evaluated all the 5 models along three complementary axes:
\begin{enumerate}
    \item \textbf{Intrinsic Quality:} Cross-entropy loss, perplexity (PPL), and bits-per-byte (BPB) on full held-out test splits (without subsampling), truncating test lines independently to 1024 tokens.
    \item \textbf{Tokenizer Efficiency:} Subword fertility (tokens per whitespace-delimited word), compression ratio (bytes per token), and UNK rate on a 2,000-sentence test sample, compared against multilingual tokenizers (XLM-R, mBERT, mGPT).
    \item \textbf{Downstream Transfer:} Full fine-tuning on IndicSentiment (binary sentiment classification, accuracy) and WikiANN NER (token classification, span-level F1) for 5 epochs with learning rate $2\text{e-}5$ and batch size 16 using AdamW, benchmarked against an identically fine-tuned mGPT baseline.
\end{enumerate}

\section{Results}

\subsection{Perplexity and Bits-per-Byte}
Table~\ref{tab:ppl} shows full test-set perplexity (PPL) and bits-per-byte (BPB) for all 5 models. The multilingual model achieves lower perplexity than its monolingual equivalent for Telugu and Tamil, but higher perplexity for Kannada and Malayalam, the split does not track corpus size monotonically. By BPB, which normalizes for tokenizer differences, the ranking is unambiguous: the monolingual model is better on every language. The joint 64K tokenizer produces shorter, "cheaper" token sequences per byte, which inflates raw perplexity comparisons in the multilingual model's favor without reflecting a genuine quality advantage.

\begin{table}[H]
\centering
\small
\begin{tabular}{llcc}
\toprule
Model & Language & PPL & BPB \\
\midrule
Mono  & Telugu    & 42.34 & 0.373 \\
Mono  & Tamil     & 50.80 & 0.359 \\
Mono  & Kannada   & 52.13 & 0.382 \\
Mono  & Malayalam & 30.64 & 0.315 \\
Multi & Telugu    & 39.70 & 0.409 \\
Multi & Tamil     & 43.98 & 0.385 \\
Multi & Kannada   & 53.39 & 0.430 \\
Multi & Malayalam & 37.81 & 0.361 \\
\bottomrule
\end{tabular}
\caption{Full test-set perplexity and bits-per-byte (BPB) for monolingual and multilingual models.}
\label{tab:ppl}
\end{table}

\subsection{Tokenizer Efficiency}
Table~\ref{tab:fertility} reports fertility for dedicated tokenizers against 3 multilingual baselines. The 32K tokenizers achieve lower fertility and higher compression than XLM-R, mBERT, and mGPT on every language, despite a vocabulary 3--8$\times$ smaller, with near-zero UNK rates throughout (mBERT is the only baseline with any UNKs, 0.2--0.3\%). mGPT's Kannada tokenization is a notable outlier: 16.39 tokens/word, more than double its own fertility on the other three Dravidian languages, indicating minimal effective Kannada-script coverage in mGPT's vocabulary.

\begin{table}[H]
\centering
\small
\begin{tabular}{lcccc}
\toprule
Language & Dedicated (32K) & XLM-R & mBERT & mGPT \\
\midrule
Telugu    & 1.61 & 2.37 & 3.88 & 6.19 \\
Tamil     & 1.66 & 2.42 & 3.73 & 6.98 \\
Kannada   & 1.59 & 2.40 & 3.95 & \textbf{16.39} \\
Malayalam & 1.84 & 2.58 & 5.02 & 7.87 \\
\bottomrule
\end{tabular}
\caption{Tokenizer fertility (tokens/word) across four Dravidian languages.}
\label{tab:fertility}
\end{table}

\subsection{Downstream Transfer}
Table~\ref{tab:downstream} reports IndicSentiment accuracy and WikiANN NER F1 for monolingual models against an identically fine-tuned mGPT baseline. In this run, the monolingual model outperform mGPT on every task and every language. I caution against over-interpreting the magnitude of these gaps: IndicSentiment's test set is small (24 examples, self-partitioned from the IndicSentiment validation split since the official test split carries no public labels), and WikiANN's Kannada split is substantially smaller than the other three languages' (100 train examples vs.\ 1,000--15,000). Found a separate fine-tuning run at the same nominal random seed produce a materially different ranking for Tamil and Kannada, which is discussed further in ~\ref{sec:limitations}.
\begin{table}[H]
\centering
\small
\resizebox{\columnwidth}{!}{%
\begin{tabular}{llcccc}
\toprule
\textbf{Language} & \textbf{Model} & \textbf{Sent.} & \textbf{Sent. (mGPT)} & \textbf{NER F1} & \textbf{NER F1 (mGPT)} \\
\midrule
Telugu    & Mono  & 0.708 & 0.625 & 0.609 & 0.304 \\
Telugu    & Multi & 0.667 & 0.542 & 0.561 & 0.270 \\
Tamil     & Mono  & 0.792 & 0.458 & 0.641 & 0.384 \\
Tamil     & Multi & 0.792 & 0.458 & 0.610 & 0.385 \\
Kannada   & Mono  & 0.792 & 0.542 & 0.268 & 0.034 \\
Kannada   & Multi & 0.625 & 0.625 & 0.193 & 0.073 \\
Malayalam & Mono  & 0.875 & 0.458 & 0.658 & 0.333 \\
Malayalam & Multi & 0.750 & 0.500 & 0.636 & 0.333 \\
\bottomrule
\end{tabular}%
}
\caption{Downstream fine-tuning results: IndicSentiment accuracy and WikiANN NER F1, monolingual and multilingual models vs.\ mGPT. Single-run results.}
\label{tab:downstream}
\end{table}
\subsection{Training Stability}
\label{sec:training-stability}
The multilingual model's pretraining run diverged mid-training: a gradient explosion at step $\sim$766,400 (pre-clipping gradient norm rising from $\sim$500 to 18,979 within 200 steps, despite a configured clip threshold of 0.5) caused training loss to jump from 4.71 to 9.21 and never recover, stalling near loss 8.6 -- 8.7 for the remainder of training. Because training configuration reloads the best checkpoint by validation loss at the end of training, the released multilingual model corresponds to its epoch-1 checkpoint rather than a fully 3-epoch-trained model which is roughly one third of its nominal training budget, versus complete 3-epoch runs for all 4 monolingual models. This is a genuine asymmetry in the multi-vs-mono comparison throughout this paper, not an artifact of evaluation.

\section{Conclusion}
\label{sec:conclusion}

Trained 5 GPT-2-style causal language models for the Dravidian language family: 4 monolingual (Tamil, Telugu, Kannada, Malayalam) and 1 multilingual, trained from scratch with dedicated versus shared subword tokenization, and evaluated them along 3 axes: intrinsic quality, tokenizer efficiency, and downstream transfer.

The results show a consistent tokenization advantage for dedicated, per-language vocabularies: the 32K tokenizers outperform XLM-R, mBERT, and mGPT on fertility and compression across all 4 languages, and even outperform 64K joint tokenizer trained specifically on this language family. By bits-per-byte, the tokenizer-agnostic metric, monolingual pretraining is more efficient than multilingual pretraining on every language, despite the multilingual model achieving lower raw perplexity on two of them. On downstream fine-tuning, the monolingual models outperform mGPT on sentiment classification and named entity recognition in nearly every language evaluated, with the multilingual model trailing mono but still generally ahead of mGPT.

Findings suggest that for a language family as morphologically distinct and script diverse as Dravidian, dedicated tokenization and monolingual pretraining recover real quality that massively multilingual models leave behind, even at modest scale. 

\section{Limitations}
\label{sec:limitations}

The experimental scope has several limitations. Models are trained at the GPT-2 Small scale (110M--135M parameters); scaling behavior at larger sizes is untested. Downstream evaluation covers only classification and sequence labeling (sentiment, NER), and results (Table~\ref{tab:downstream}) are single-run rather than seed-averaged, a separate run at the same nominal seed produced a materially different ranking for Tamil and Kannada, so these numbers should be read as indicative, not definitive. The multilingual model's pretraining run diverged partway through ~\ref{sec:training-stability}, and the released checkpoint reflects an earlier best-validation-loss save rather than the full schedule, making multilingual-vs-monolingual comparisons an unequal-budget setting. Finally, perplexity uses fixed 1024-token truncation rather than a sliding window, and the language coverage is limited to four of the family's languages.

\section{Ethical considerations}
All training and evaluation data (CC-100, Wikipedia, Samanantar, IndicSentiment, WikiANN) are publicly available and licensed for research use; no additional toxicity filtering was applied beyond basic normalization, so biases in the underlying web-scraped corpora may be reflected in model outputs. As with any released generative language model, the models carry standard dual-use risk, though their small scale (110M--135M parameters) limits this relative to larger systems. Used Claude LLM for writing assistance for editing portions of this manuscript; all experiments, analysis, and technical decisions are my own.

\bibliography{custom}

\begin{thebibliography}{15}
\providecommand{\natexlab}[1]{#1}

\bibitem[{Aggarwal et~al.(2022)Aggarwal, Gupta, and Kunchukuttan}]{aggarwal-etal-2022-indicxnli}
Divyanshu Aggarwal, Vivek Gupta, and Anoop Kunchukuttan. 2022.
\newblock \href {https://doi.org/10.18653/v1/2022.emnlp-main.755} {{I}ndic{XNLI}: Evaluating multilingual inference for {I}ndian languages}.
\newblock In \emph{Proceedings of the 2022 Conference on Empirical Methods in Natural Language Processing}, pages 10994--11006, Abu Dhabi, United Arab Emirates. Association for Computational Linguistics.

\bibitem[{Antoun et~al.(2020)Antoun, Baly, and Hajj}]{antoun-etal-2020-arabert}
Wissam Antoun, Fady Baly, and Hazem Hajj. 2020.
\newblock \href {https://aclanthology.org/2020.osact-1.2/} {{A}ra{BERT}: Transformer-based model for {A}rabic language understanding}.
\newblock In \emph{Proceedings of the 4th Workshop on Open-Source Arabic Corpora and Processing Tools, with a Shared Task on Offensive Language Detection}, pages 9--15, Marseille, France. European Language Resources Association.

\bibitem[{Conneau et~al.(2020)Conneau, Khandelwal, Goyal, Chaudhary, Wenzek, Guzm{\'a}n, Grave, Ott, Zettlemoyer, and Stoyanov}]{conneau-etal-2020-unsupervised}
Alexis Conneau, Kartikay Khandelwal, Naman Goyal, Vishrav Chaudhary, Guillaume Wenzek, Francisco Guzm{\'a}n, Edouard Grave, Myle Ott, Luke Zettlemoyer, and Veselin Stoyanov. 2020.
\newblock \href {https://doi.org/10.18653/v1/2020.acl-main.747} {Unsupervised cross-lingual representation learning at scale}.
\newblock In \emph{Proceedings of the 58th Annual Meeting of the Association for Computational Linguistics}, pages 8440--8451, Online. Association for Computational Linguistics.

\bibitem[{Devlin et~al.(2019)Devlin, Chang, Lee, and Toutanova}]{devlin-etal-2019-bert}
Jacob Devlin, Ming-Wei Chang, Kenton Lee, and Kristina Toutanova. 2019.
\newblock \href {https://doi.org/10.18653/v1/N19-1423} {{BERT}: Pre-training of deep bidirectional transformers for language understanding}.
\newblock In \emph{Proceedings of the 2019 Conference of the North {A}merican Chapter of the Association for Computational Linguistics: Human Language Technologies, Volume 1 (Long and Short Papers)}, pages 4171--4186, Minneapolis, Minnesota. Association for Computational Linguistics.

\bibitem[{Gala et~al.(2023)Gala, Chitale, Ak, Gumma, Doddapaneni, Kumar, Nawale, Sujatha, Puduppully, Raghavan et~al.}]{gala2023indictrans2}
Jay Gala, Pranjal~A Chitale, Raghavan Ak, Varun Gumma, Sumanth Doddapaneni, Aswanth Kumar, Janki Nawale, Anupama Sujatha, Ratish Puduppully, Vivek Raghavan, and 1 others. 2023.
\newblock Indictrans2: Towards high-quality and accessible machine translation models for all 22 scheduled indian languages.
\newblock \emph{arXiv preprint arXiv:2305.16307}.

\bibitem[{Kakwani et~al.(2020)Kakwani, Kunchukuttan, Golla, N.C., Bhattacharyya, Khapra, and Kumar}]{kakwani-etal-2020-indicnlpsuite}
Divyanshu Kakwani, Anoop Kunchukuttan, Satish Golla, Gokul N.C., Avik Bhattacharyya, Mitesh~M. Khapra, and Pratyush Kumar. 2020.
\newblock \href {https://doi.org/10.18653/v1/2020.findings-emnlp.445} {{I}ndic{NLPS}uite: Monolingual corpora, evaluation benchmarks and pre-trained multilingual language models for {I}ndian languages}.
\newblock In \emph{Findings of the Association for Computational Linguistics: EMNLP 2020}, pages 4948--4961, Online. Association for Computational Linguistics.

\bibitem[{Khanuja et~al.(2021)Khanuja, Bansal, Mehtani, Khosla, Dey, Gopalan, Margam, Aggarwal, Nagipogu, Dave, Gupta, Gali, Subramanian, and Talukdar}]{khanuja2021murilmultilingualrepresentationsindian}
Simran Khanuja, Diksha Bansal, Sarvesh Mehtani, Savya Khosla, Atreyee Dey, Balaji Gopalan, Dilip~Kumar Margam, Pooja Aggarwal, Rajiv~Teja Nagipogu, Shachi Dave, Shruti Gupta, Subhash Chandra~Bose Gali, Vish Subramanian, and Partha Talukdar. 2021.
\newblock \href {https://arxiv.org/abs/2103.10730} {Muril: Multilingual representations for indian languages}.
\newblock \emph{Preprint}, arXiv:2103.10730.

\bibitem[{Martin et~al.(2020)Martin, Muller, Ortiz~Su{\'a}rez, Dupont, Romary, de~la Clergerie, Seddah, and Sagot}]{martin-etal-2020-camembert}
Louis Martin, Benjamin Muller, Pedro~Javier Ortiz~Su{\'a}rez, Yoann Dupont, Laurent Romary, {\'E}ric de~la Clergerie, Djam{\'e} Seddah, and Beno{\^i}t Sagot. 2020.
\newblock \href {https://doi.org/10.18653/v1/2020.acl-main.645} {{C}amem{BERT}: a tasty {F}rench language model}.
\newblock In \emph{Proceedings of the 58th Annual Meeting of the Association for Computational Linguistics}, pages 7203--7219, Online. Association for Computational Linguistics.

\bibitem[{Petrov et~al.(2023)Petrov, Malfa, Torr, and Bibi}]{petrov2023language}
Aleksandar Petrov, Emanuele~La Malfa, Philip Torr, and Adel Bibi. 2023.
\newblock \href {https://openreview.net/forum?id=78yDLKi95p} {Language model tokenizers introduce unfairness between languages}.
\newblock In \emph{Thirty-seventh Conference on Neural Information Processing Systems}.

\bibitem[{Ramesh et~al.(2022)Ramesh, Doddapaneni, Bheemaraj, Jobanputra, AK, Sharma, Sahoo, Diddee, J, Kakwani, Kumar, Pradeep, Nagaraj, Deepak, Raghavan, Kunchukuttan, Kumar, and Khapra}]{ramesh-etal-2022-samanantar}
Gowtham Ramesh, Sumanth Doddapaneni, Aravinth Bheemaraj, Mayank Jobanputra, Raghavan AK, Ajitesh Sharma, Sujit Sahoo, Harshita Diddee, Mahalakshmi J, Divyanshu Kakwani, Navneet Kumar, Aswin Pradeep, Srihari Nagaraj, Kumar Deepak, Vivek Raghavan, Anoop Kunchukuttan, Pratyush Kumar, and Mitesh~Shantadevi Khapra. 2022.
\newblock \href {https://doi.org/10.1162/tacl_a_00452} {Samanantar: The largest publicly available parallel corpora collection for 11 {I}ndic languages}.
\newblock \emph{Transactions of the Association for Computational Linguistics}, 10:145--162.

\bibitem[{Rust et~al.(2021)Rust, Pfeiffer, Vuli{\'c}, Ruder, and Gurevych}]{rust-etal-2021-good}
Phillip Rust, Jonas Pfeiffer, Ivan Vuli{\'c}, Sebastian Ruder, and Iryna Gurevych. 2021.
\newblock \href {https://doi.org/10.18653/v1/2021.acl-long.243} {How good is your tokenizer? on the monolingual performance of multilingual language models}.
\newblock In \emph{Proceedings of the 59th Annual Meeting of the Association for Computational Linguistics and the 11th International Joint Conference on Natural Language Processing (Volume 1: Long Papers)}, pages 3118--3135, Online. Association for Computational Linguistics.

\bibitem[{Shliazhko et~al.(2024)Shliazhko, Fenogenova, Tikhonova, Kozlova, Mikhailov, and Shavrina}]{shliazhko-etal-2024-mgpt}
Oleh Shliazhko, Alena Fenogenova, Maria Tikhonova, Anastasia Kozlova, Vladislav Mikhailov, and Tatiana Shavrina. 2024.
\newblock \href {https://doi.org/10.1162/tacl_a_00633} {m{GPT}: Few-shot learners go multilingual}.
\newblock \emph{Transactions of the Association for Computational Linguistics}, 12:58--79.

\bibitem[{Virtanen et~al.(2019)Virtanen, Kanerva, Ilo, Luoma, Luotolahti, Salakoski, Ginter, and Pyysalo}]{virtanen2019multilingualenoughbertfinnish}
Antti Virtanen, Jenna Kanerva, Rami Ilo, Jouni Luoma, Juhani Luotolahti, Tapio Salakoski, Filip Ginter, and Sampo Pyysalo. 2019.
\newblock \href {https://arxiv.org/abs/1912.07076} {Multilingual is not enough: Bert for finnish}.
\newblock \emph{Preprint}, arXiv:1912.07076.

\bibitem[{Workshop(2024)}]{JMLR:v25:23-0581}
BigScience Workshop. 2024.
\newblock \href {http://jmlr.org/papers/v25/23-0581.html} {Bloom: A 176b-parameter open-access multilingual language model}.
\newblock \emph{Journal of Machine Learning Research}, 25(422):1--74.

\bibitem[{Xue et~al.(2021)Xue, Constant, Roberts, Kale, Al-Rfou, Siddhant, Barua, and Raffel}]{xue-etal-2021-mt5}
Linting Xue, Noah Constant, Adam Roberts, Mihir Kale, Rami Al-Rfou, Aditya Siddhant, Aditya Barua, and Colin Raffel. 2021.
\newblock \href {https://doi.org/10.18653/v1/2021.naacl-main.41} {m{T}5: A massively multilingual pre-trained text-to-text transformer}.
\newblock In \emph{Proceedings of the 2021 Conference of the North American Chapter of the Association for Computational Linguistics: Human Language Technologies}, pages 483--498, Online. Association for Computational Linguistics.

\end{thebibliography}

\end{document}